\documentclass[10pt,twocolumn,letterpaper]{article}

\usepackage{cvpr}              

\usepackage[dvipsnames]{xcolor}

\usepackage{multirow}

\usepackage{algorithm,algcompatible,amsmath,}
\usepackage{makecell}
\algdef{SE}[SUBALG]{Indent}{EndIndent}{}{\algorithmicend\ }%
\algnewcommand\INPUT{\item[\textbf{Input:}]}%
\algnewcommand\OUTPUT{\item[\textbf{Output:}]}%
\algnewcommand\PARAM{\item[\textbf{Optimizable Parameter:}]}%

\algdef{SE}[SUBALG]{Indent}{EndIndent}{}{\algorithmicend\ }%
\algtext*{Indent}
\algtext*{EndIndent}

\definecolor{cvprblue}{rgb}{0.21,0.49,0.74}
\usepackage[pagebackref,breaklinks,colorlinks,allcolors=cvprblue]{hyperref}

\def\paperID{8591} 
\def\confName{CVPR}
\def\confYear{2025}

\def\titlePrefix{ActiveGAMER}
\title{\titlePrefix{}: Active GAussian Mapping through Efficient Rendering}

\author{
Liyan Chen 
    \thanks{Equal contribution} 
    \thanks{Work done as an intern at OPPO US Research Center} 
    \textsuperscript{,1,2}
    \hspace{15pt}
Huangying Zhan 
    \footnotemark[1]
    \thanks{Corresponding author (zhanhuangying.work@gmail.com)}
    \textsuperscript{,1} 
    \hspace{15pt}
Kevin Chen
    \textsuperscript{1} 
    \hspace{15pt}
Xiangyu Xu
    \textsuperscript{1} 
    \hspace{10pt}
    \\
Qingan Yan
    \textsuperscript{1} 
    \hspace{10pt}
Changjiang Cai
    \textsuperscript{1} 
    \hspace{10pt}
Yi Xu
    \textsuperscript{1} 
    \and 
\textsuperscript{1}
    OPPO US Research Center \and
\textsuperscript{2}
    Stevens Institute of Technology
}

\begin{document}

\maketitle
\begin{abstract}
We introduce \textit{\titlePrefix{}}, an active mapping system that utilizes 3D Gaussian Splatting (3DGS) to achieve high-quality scene mapping and efficient exploration. 
Unlike recent NeRF-based methods, which are computationally demanding and limit mapping performance, our approach leverages the efficient rendering capabilities of 3DGS to enable effective and efficient exploration in complex environments.
The core of our system is a rendering-based information gain module that identifies the most informative viewpoints for next-best-view planning, enhancing both geometric and photometric reconstruction accuracy. 
\textit{\titlePrefix{}} also integrates a carefully balanced framework, combining coarse-to-fine exploration, post-refinement, and a global-local keyframe selection strategy to maximize reconstruction completeness and fidelity.
Our system autonomously explores and reconstructs environments with state-of-the-art geometric and photometric accuracy and completeness, significantly surpassing existing approaches in both aspects. 
Extensive evaluations on benchmark datasets such as Replica and MP3D highlight \textit{\titlePrefix{}}'s effectiveness in active mapping tasks.
\end{abstract}
\section{Introduction}
\label{sec:intro}

\begin{figure}[t!]
        \centering
		\includegraphics[width=1.0\columnwidth]    {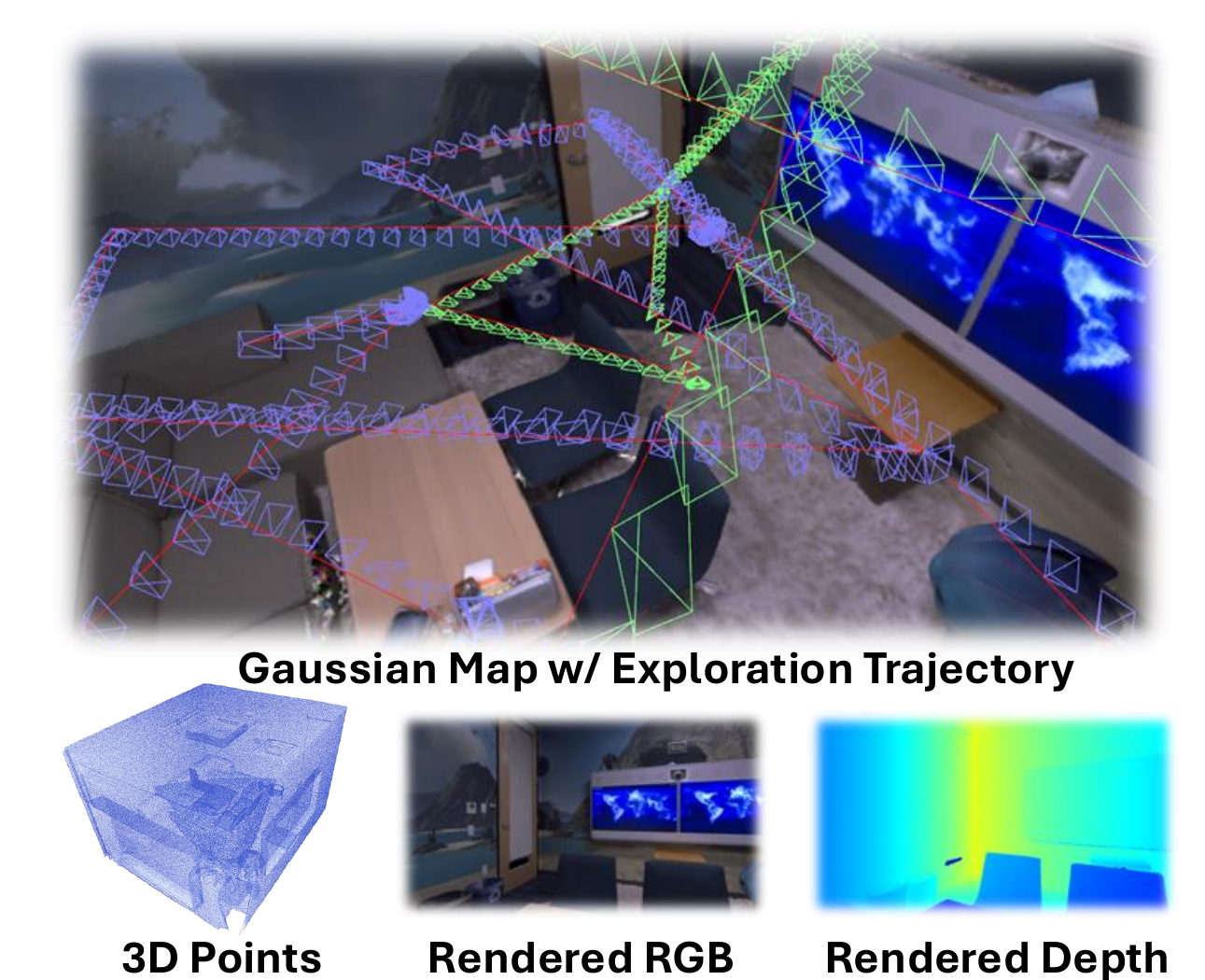}
		\caption{
            \textit{\titlePrefix{}} is built on a Gaussian Map backbone and autonomously performs \textcolor[HTML]{8183FB}{coarse}-to-\textcolor[HTML]{85FC86}{fine} exploration to optimize geometric accuracy and photometric fidelity.
            }
		\label{fig:teaser}
		\vspace{-15pt}
\end{figure}

In computer vision, the ability to generate detailed 3D reconstructions from 2D images or videos has seen tremendous progress, enabling real-time, incremental 3D modeling as new visual data is assimilated. 
This process, often powered by Simultaneous Localization and Mapping (SLAM), plays a crucial role in robotic applications, where it supports tasks such as planning and navigation. 
When combined, these functions define Active SLAM, a framework that integrates localization, mapping, planning, and navigation to enable autonomous exploration.

This paper focuses on a subproblem of Active SLAM known as Active Reconstruction, where localization is assumed to be known, allowing the system to prioritize high-quality and complete scene reconstruction. 
We explore this task by introducing a novel approach based on 3D Gaussian Splatting (3DGS), which provides an efficient radiance field representation optimized for active exploration. 

Radiance fields, particularly implicit models like NeRFs, have shown promising results in applications such as 3D object reconstruction \cite{park2019deepsdf}, novel view synthesis \cite{mildenhall2021nerf, zhang2020nerf++, yu2021pixelnerf, pumarola2021d}, and surface reconstruction \cite{li2022bnv, azinovic2022neural}. 
However, NeRF’s high computational demands hinder real-time applicability, especially in active vision tasks. 
To address this, recent efforts have explored hybrid neural representations, which incorporate both implicit and explicit components to enhance rendering efficiency \cite{SunSC22dvgo, mueller2022instant}.

On the other hand, 3D Gaussian Splatting \cite{kerbl20233dgs} is a recent and efficient radiance field representation that leverages Gaussian primitives in 3D space. 
Unlike NeRFs, which rely on dense, computationally intensive sampling, 3DGS uses a sparse set of Gaussian ellipsoids to approximate both geometry and color information. 
This representation allows for rapid rendering by projecting each Gaussian onto the image plane, where they are blended using alpha compositing. 
Due to its efficient design, 3DGS achieves real-time rendering speeds, making it particularly suitable for dynamic applications like SLAM and active vision, where computational resources and time are limited. 
Recent work \cite{huang20242dgs, yu2024mipsplatting} has shown that 3DGS maintains high visual fidelity while reducing the computational overhead, enabling its use in scenarios that demand fast, high-quality 3D reconstructions.

Despite these advances, integrating radiance fields into active vision remains challenging. 
While several studies have explored active reconstruction and path planning with NeRFs \cite{adamkiewicz2022vision, ran2022neurar, lee2022uncertainty, pan2022activenerf, zhan2022activermap}, these methods often focusing on geometric reconstruction tasks, especially surface reconstruction. 
Improving photometric reconstruction (rendering) is often ignored with the use of NeRF-based methods due to its slow rendering speeds. 

To address existing limitations and enhance photometric reconstruction, we introduce \textit{\titlePrefix{}}, an active mapping system that leverages the real-time rendering capabilities of 3D Gaussian Splatting. 
Our system enables unrestricted 6DoF movement, allowing flexible exploration and high-quality scene reconstruction even in complex environments. 
The key contributions of our work include:
\begin{itemize} 
\setlength{\itemsep}{0pt} 
\setlength{\parskip}{0pt} 
\setlength{\parsep}{0pt} 
    \item An advanced active mapping system based on 3DGS, allowing real-time, unrestricted 6DoF exploration. 
    \item A rendering-based information gain module that efficiently identifies the most informative viewpoints for next-best-view planning. 
    \item A carefully designed system that balances exploration efficiency and reconstruction accuracy for both geometric and photometric reconstruction, through strategies like coarse-to-fine exploration, post-refinement, and global-local keyframe selection.
    \item Achieves state-of-the-art performance in active reconstruction, enhancing both geometric accuracy and rendering fidelity over existing approaches.
\end{itemize}

\section{Related Work} 
\label{sec:rel_work}

\begin{figure*}[th!]
        \centering
		\includegraphics[width=1.0\textwidth]{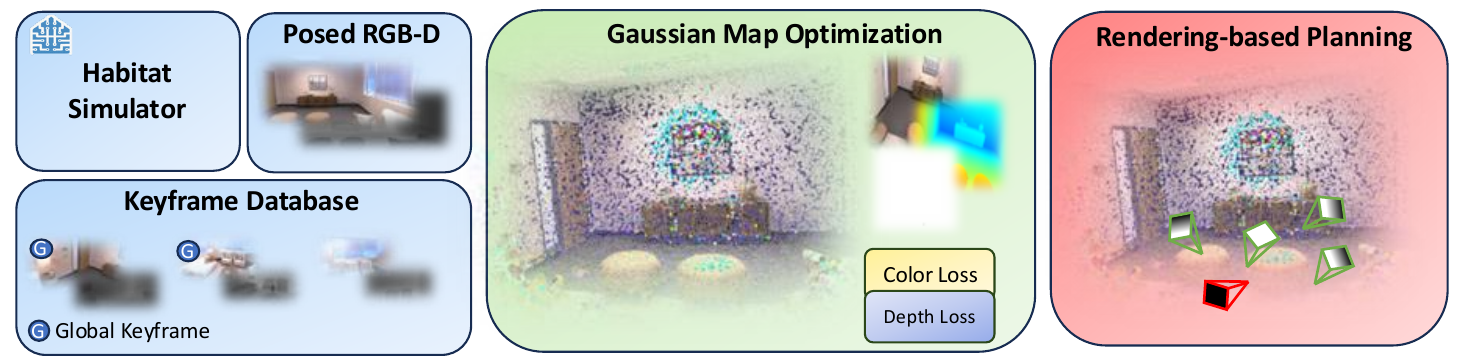}
		\caption{
            \textbf{\titlePrefix{} Framework.}
            At each keyframe step, HabitatSim \cite{savva2019habitat} generates posed RGB-D images, which are stored in a keyframe database, with certain frames designated as Global Keyframes. 
            These observations are used to update a Gaussian Map comprising a collection of 3D Gaussians. 
            Map optimization is achieved by minimizing color and depth rendering losses, based on the rendered RGB-D images and silhouette masks. 
            Using the up-to-date Gaussian Map, rendering-based planning evaluates the information gain across sampled candidate viewpoints and choose the one with highest information gain as the next-best-view. 
                }
		\label{fig:framework}
\end{figure*}

\paragraph{Active Reconstruction}
Autonomous robotics relies on foundational capabilities such as localization, mapping, planning, and motion control \cite{siegwart2011introduction}.
These capabilities have spurred advancements in various areas, including visual odometry \cite{scaramuzza2011visual, zhan2020visual}, monocular depth estimation \cite{eigen2014depth,  zhan2018unsupervised, bian2019unsupervised}, multi-view stereo \cite{sun2003stereo, hirschmuller2005accurate, seitz2006comparison, yao2018mvsnet, liu2022planemvs, cai2023riav, chen2023learning}, structure-from-motion (SfM) \cite{schonberger2016structure}, and path planning \cite{Hart1968, lavalle2001rrt, kuffner2000rrtconnect}.
SLAM systems have also evolved significantly, facilitating simultaneous localization and mapping for autonomous navigation \cite{thrun2002probabilistic, durrant2006simultaneous, davison2007monoslam, cadena2016past, wang2023edi}.
Active SLAM combines these approaches to enable autonomous localization, mapping, and planning, aiming to reduce uncertainty in the robot’s environment representation \cite{davison2002simultaneous}.
Comprehensive reviews of active SLAM can be found in recent survey papers \cite{cadena2016past, lluvia2021active, placed2022survey}.
Our focus is on active reconstruction, a research area closely related to active SLAM but oriented towards achieving complete and accurate 3D representations.
Active reconstruction is often formulated as an exploration problem, where the objective is to determine the most informative viewpoints for capturing detailed scene representations \cite{thrun1991active, feder1999adaptive, bourgault2002information, makarenko2002experiment, stachniss2004exploration, newman2003autonomous, stachniss2009robotic}.
This approach has been widely applied in scene and object reconstruction from multiple viewpoints, with methods developed to handle occlusions, optimize viewpoint selection, and maximize information gain \cite{connolly1985determination, maver1993occlusions, pito1999solution, kriegel2015efficient, isler2016information, delmerico2018comparison, peralta2020next}.

\paragraph{Radiance Fields}
Radiance Field representations have become a cornerstone in 3D scene modeling, enabling continuous and high-quality representations of complex environments.
Neural Radiance Fields (NeRFs) \cite{mildenhall2021nerf} pioneered this field by modeling scenes as continuous radiance fields using multi-layer perceptrons (MLPs), achieving impressive results in applications such as novel view rendering \cite{mildenhall2021nerf, zhang2020nerf++, yu2021pixelnerf, pumarola2021d}, object and surface reconstruction \cite{park2019deepsdf, mildenhall2021nerf, li2022bnv, azinovic2022neural}, and generative modeling \cite{schwarz2020graf, niemeyer2021giraffe}, and Structure-from-Motion \cite{wang2021nerf--, lin2021barf, chng2022gaussian}.
While NeRFs have set a high standard for rendering quality, they are computationally intensive, making real-time applications challenging.

Recent advancements in 3D Gaussian Splatting \cite{kerbl20233dgs} provide an alternative approach by representing scenes through a set of Gaussian primitives in 3D space, which can be rendered efficiently with lower computational overhead.
3D Gaussian Splatting has shown promise in achieving real-time rendering, making it suitable for dynamic applications like SLAM \cite{keetha2024splatam, matsuki2024gaussian}.
Together, NeRFs and 3D Gaussian Splatting represent complementary approaches to scene representation, each with unique strengths and trade-offs in terms of rendering quality, computational efficiency, and suitability for active vision tasks.

\paragraph{Active Radiance Fields}
Building on advancements in NeRF and 3D Gaussian Splatting, recent research has investigated the potential of these representations in active vision, particularly for autonomous scene exploration and mapping.
Active radiance fields leverage the strengths of NeRF and 3DGS for tasks requiring continuous exploration and decision-making in 3D environments. 

NeRF-based approaches have been applied to path planning \cite{adamkiewicz2022vision} and active reconstruction via next-best-view strategies \cite{ran2022neurar, lee2022uncertainty, pan2022activenerf}, though they are often limited by NeRF’s high computational demands, which restricts real-time applications \cite{kuang2024active}.
To overcome these limitations, hybrid models such as ActiveRMAP \cite{zhan2022activermap} integrate implicit and explicit representations for improved efficiency. 
However, many NeRF-based methods still restrict camera motion to constrainted spaces, reducing flexibility in complex 3D environments. 
NARUTO \cite{feng2024naruto} addresses this by introducing an active neural mapping system with 6DoF movement in unrestricted spaces, while \cite{kuang2024active} integrates Voronoi planning to scale exploration across larger environments.

3DGS  offers a faster alternative, making real-time mapping and exploration more feasible. Concurrent works like ActiveSplat \cite{li2024activesplat} utilize a hybrid map with topological abstractions for efficient planning, and AG-SLAM \cite{jiang2024agslam} incorporates 3DGS with Fisher Information to balance exploration and localization in complex environments.
In this work, we propose an Active Gaussian Mapping method that leverages the real-time rendering capability of 3DGS to enable effective planning and exploration.
\section{\titlePrefix{}: Active Gaussian Mapping}
\label{sec:method}


In this section, we present \textit{\titlePrefix{}} (\cref{fig:framework}), a pioneering 3D Gaussian Splatting framework for active reconstruction that incorporates rendering-based planning.
Our approach begins with the Gaussian Splatting mapping module, an efficient representation for real-time, high-fidelity geometric and photometric reconstruction.
We utilize SplaTAM \cite{keetha2024splatam} as the mapping backbone, as detailed in \cref{sec:method:gaussian_map}, establishing a foundation for dense reconstruction using Gaussian Maps.
Building upon this, \cref{sec:method:render_planning} introduces our rendering-based planning module for goal-directed searching and path planning.
To address the limitations of the map update process in SplaTAM, we propose an enhanced update strategy in \cref{sec:method:keyframe_selection}, incorporating a global-local keyframe selection strategy.
This module leverages rendering-based information, ensuring seamless integration into existing incremental 3DGS to improve mapping robustness.
We then present a post-refinement module that further improves the photometric reconstruction in \cref{sec:post_refinement}.
Finally, we conclude this section with an overview of the Active Gaussian Mapping process in \cref{sec:method:active_gs}.

\subsection{Gaussian Mapping}
\label{sec:method:gaussian_map}
\paragraph{Gaussian Splatting}
Recent advancements have established 3D Gaussian Splatting as both expressive and efficient.
These representations effectively encode a scene's appearance and 3D geometry as a collection of 3D Gaussians, $\textbf{G} = \{ \mathbf{G}_i \}_{i=1}^N$, 
enabling real-time rendering into high-fidelity color and depth images.
A series of prior works, including \cite{matsuki2024gaussian, keetha2024splatam}, have demonstrated the effectiveness of 3DGS in 3D reconstruction.
Given a stream of RGB-D images, dense mapping is done by optimizing the 3D Gaussian representation through rendering supervision.

Rather than using the comprehensive 3DGS representation originally proposed in \cite{kerbl20233dgs}, we adopt the simplified Gaussian Mapping suggested in SplaTAM \cite{keetha2024splatam}.
This simplified approach utilizes only view-independent color and isotropic Gaussians, reducing the number of parameters needed for each Gaussian.
The parameters for each Gaussian include color $\textbf{c}$, center position $\boldsymbol{\mu}$, radius $r$, and opacity $o$.
A 3D point $\textbf{x}$ is influenced by all Gaussians based on a standard Gaussian function weighted by its opacity:
\begin{equation}
    f(\mathbf{x}) = o \exp \left( -\frac{|\mathbf{x} - \boldsymbol{\mu}|^2}{2r^2} \right).
\end{equation} \label{eqn:3d_gaussian}
\paragraph{Real-time Rendering}
A key strength of 3DGS is its real-time rendering capability, enabling high-fidelity color and depth image generation from arbitrary camera poses.
Leveraging Gaussian Maps, 3DGS can efficiently render scenes by projecting 3D Gaussians into 2D pixel space.
Following the approach proposed in \cite{kerbl20233dgs}, we achieve efficient rendering by transforming all 3D Gaussians to camera space, sorting them front-to-back, and projecting them onto the image plane.
Each Gaussian is then splatted in 2D, where color and transparency are composited using alpha-blending.
The color at a pixel, $\mathbf{p} = (u, v)$, is defined as:
\begin{equation}
C(\mathbf{p}) = \sum_{i=1}^{n} c_i f_i(\mathbf{p}) \prod_{j=1}^{i-1} \left( 1 - f_j(\mathbf{p}) \right),
\end{equation}
where $f_i(\mathbf{p})$ is derived from the Gaussian’s position and size in 2D pixel space, computed as:
\begin{equation}
\boldsymbol{\mu}^{\text{2D}} = K \frac{T_t \boldsymbol{\mu}}{d}, 
\quad r^{\text{2D}} = \frac{f r}{d}, 
\quad \text{where} \quad d = (T_t \boldsymbol{\mu})_z.
\end{equation}
Here, $K$ represents the camera intrinsics, including focal length $f$ and principal point, while $T_t$ encodes the camera’s extrinsics, capturing its rotation and translation in world space at time $t$.
The variable $d$ represents the depth of the Gaussian in camera coordinates.

Similarly, the depth map is rendered as:
\begin{equation}
D(\mathbf{p}) = \sum_{i=1}^{n} d_i f_i(\mathbf{p}) \prod_{j=1}^{i-1} \left( 1 - f_j(\mathbf{p}) \right).
\end{equation}

We also generate a silhouette mask to indicate if a pixel contains information from the Gaussian map:
\begin{equation}
S(\mathbf{p}) = \sum_{i=1}^{n} f_i(\mathbf{p}) \prod_{j=1}^{i-1} \left( 1 - f_j(\mathbf{p}) \right).
\end{equation}

\paragraph{Optimization}
The differentiable nature of this rendering process enables end-to-end optimization, where gradients are calculated directly from the discrepancy between rendered images and RGB-D inputs.
These gradients drive updates to each Gaussian’s parameters through minimization of the following loss:
\begin{equation}
L = \sum_{\mathbf{p}} \left( S(\mathbf{p}) > 0.99 \right) \left( L_1(D(\mathbf{p})) + 0.5 L_1(C(\mathbf{p})) \right).
\end{equation}
By thresholding the silhouette mask, we selectively optimize pixels in regions with high-quality visibility, enhancing the stability of the reconstruction.

\paragraph{Gaussian Densification}
Gaussian Densification is designed to adaptively add new 3D Gaussians in response to incoming data.
Using the known camera pose and depth measurements, we compute a densification mask to determine where additional Gaussians are needed, avoiding redundant creation in regions already well-represented by the current model.
The densification mask is defined as:
\begin{align}
M(\mathbf{p}) &= \left( S(\mathbf{p}) < 0.5 \right) + \notag \\
& \left( D_{\text{GT}}(\mathbf{p}) < D(\mathbf{p}) \right) \left( L_1(D(\mathbf{p})) > \lambda \text{MDE} \right).
\end{align}
This mask identifies areas where:
(1) the density of Gaussians is low ($S < 0.5$), and
(2) new Gaussians are required to refine the geometry, as indicated by a depth error exceeding a threshold of $\lambda = 50$ times the median depth error (MDE).

\subsection{Rendering-based Planning} \label{sec:method:render_planning}

In \cref{sec:method:gaussian_map}, we introduce a Gaussian Mapping method using known camera parameters and incremental RGB-D observations.
Typically, mapping is conducted in a passive manner, where the capture trajectory is controlled manually.
However, passive capture does not ensure a complete, high-fidelity reconstruction of the scene.
In this section, we propose an active mapping system that leverages the Gaussian Map, specifically exploiting its efficient rendering capabilities—advantages that are unattainable with neural radiance fields \cite{yan2023active, feng2024naruto}. 
We propose a rendering-based information gain to achieve active exploration.

Our approach involves a two-stage active mapping strategy.
First, a coarse-to-fine exploration stage reconstructs the scene as comprehensively as possible.
Subsequently, a post-refinement stage further enhances the Gaussian Mapping representation for rendering purpose once the exploration phase is complete.


In the exploration stage, our goal is to incrementally determine the next-best-view (NBV) to enhance the completeness of the Gaussian Map.
Leveraging the real-time rendering capability of the Gaussian Map, we can extensively generate NBV candidates across the environment and evaluate the information gain for each candidate.
We first present our exploration information gain formulation, followed by a mechanism for efficiently maintaining a candidate pool to enable rapid information gain evaluation.
After that, we present a coarse-to-fine exploration strategy that further accelerates exploration efficiency.
Finally, we introduce a local path planner introduced in \cite{feng2024naruto}.

\paragraph{Exploration Information Gain}
Given a candidate camera pose, we compute its exploration information gain, $\mathcal{I}$, based on the rendered silhouette mask $S$ with respect to the up-to-date Gaussian Map.
The number of missing pixels in the rendered silhouette mask, $N_{S_i}$, quantifies the information gain for the candidate viewpoint.
We further incorporate a motion cost represented by the $L_2$ distance between the current location $T_{t,x}$ and the candidate location $T_{i,x}$, aiming to reduce overall travel distance during exploration.
That is, we prefer a candidate closer to the current camera pose when two candidates yield the same information gain.
The distance-weighted information gain is formulated as:
\begin{equation}
\mathcal{I} = (1 - \sigma(l_i)) \cdot \sigma(\log(N_{S_i})), \label{eqn:explore_ig}
\end{equation}
where $\sigma(\cdot)$ is the Softmax function, which normalizes both the distance weighting and $N_{S_i}$ in a relative form;
$l_i = \| T_{i,x} - T_{t,x} \|_2$ represents the travel distance, 
and $N_{S_i} = \sum_{\mathbf{p}} \mathbb{I}(S_i(\mathbf{p}) = 0)$ counts the pixels in the silhouette mask that are zero.
We select the candidate viewpoint with the highest $\mathcal{I}$ as the goal pose $T_g$.

\paragraph{Exploration Candidate Pool}
While Gaussian Maps provide real-time rendering capabilities, extensively sampling viewpoints within the scene and evaluating their information gain remains computationally expensive.
To manage this, we maintain an Exploration Candidate Pool, allowing us to add new sampled candidate viewpoints, update their associated information gain, and remove them from the pool when necessary.

Adding candidates from the current observation requires careful consideration, as redundant candidates with overlapping view frustums can occur.
To ensure uniform spatial sampling, we incrementally update an occupancy grid for each incoming observation.
We refer to the free space within this grid as the Exploration Map, which we use to sample exploration candidates.
Rather than sampling the entire exploration map each time, we identify newly added free space voxels by comparing the latest Exploration Map to its previous state and sample candidates only from these new voxels.
Candidate positions $T_{e,x}$ are generated from every $v_1$ meter, with evenly distributed $v_2$ viewing directions based on the Fibonacci lattice.
Once candidate poses $T_e$ are added to the Exploration Pool, we evaluate all candidates using \cref{eqn:explore_ig} and update their information gain, $\mathcal{I}_i$.
For candidates with associated $N_{S_i}$ values below 0.5\% of the total pixel count, we consider the viewpoint to be well observed and remove the candidate from the pool to prevent its overpopulation.

\paragraph{Coarse-to-fine Exploration}
Oversampling candidates enhances scene coverage and optimizes next-best-view selection but increases evaluation time.
To improve exploration efficiency, we introduce a coarse-to-fine strategy, which initially covers the environment with minimal candidates and then refines the reconstruction with finer exploration.
In the coarse stage, we sample new free voxels at specific heights, such as a single 2D plane, using larger spatial steps ($v_1 = 1$) and fewer viewing directions ($v_2 = 5$).
The fine stage increases sampling density, employing multiple height levels, smaller steps ($v_1=0.5$), and more viewing directions ($v_2=15$) to refine the reconstruction. 
Initially, sampling is restricted to newly added free space; at the start of the fine stage, we sample the entire free space, quickly discarding redundant candidates as most regions have already been observed.
This strategy effectively balances exploration speed and comprehensive scene coverage.

\paragraph{Local Path Planning}

Once the goal pose is identified, our path planning module initiates to generate a viable path from the current pose, $T_t$, to the goal pose, $T_g$.
For this, we use a sampling-based path planning approach similar to Rapid-exploration Random Tree (RRT) \cite{lavalle2001rrt}, with the Exploration Map as the basis.
Executing standard RRT in a large-scale 3D environment can be highly time-intensive.
To address this, we adopt the efficient RRT implementation proposed by \cite{feng2024naruto}, and we generate a collision-free path connecting the current and goal locations by sampling within the free space.
Additionally, we implement a rotation planning module to ensure a smooth transition from the current orientation to the goal orientation.


\begin{algorithm}  [t]
    \caption{\titlePrefix{}}
  \begin{algorithmic}[1]
    \STATE \textbf{Initialization} 
        Camera Pose~$T_0$; 
        Gaussian Map~$\textbf{G}_0$; Exploration Map~$\textbf{M}_{e,0}$;
        Observations~$\{O\}_{i=0}^{0}$; 
        Exploration Pool~$P_e$;
        \textbf{PLAN\_REQUIRED}~=~True;
        $t=0$
        \STATE \textcolor[HTML]{979759}{\# EXPLORATION}
        \WHILE {$P_e \neq \emptyset \lor t = 0$}
                \STATE $t = t + 1$
                \STATE \textcolor[HTML]{6F9759}{\# Update Database in keyframe steps}
                \STATE \textbf{Observation}: acquire a new observation $O_t$
                \STATE \textbf{Update Database}: $\{O\}_{i=0}^{t} \leftarrow \{O\}_{i=0}^{t-1}$

                
                \STATE \textcolor[HTML]{6F9759}{\# Update Mapping Models}
                \STATE \textbf{Map Update}: 
                $\textbf{G}_{t} \leftarrow \textbf{G}_{t-1}$,
                $\textbf{M}_{e,t} \leftarrow \textbf{M}_{e,t-1}$,

                \STATE \textcolor[HTML]{6F9759}{\# Update Exploration Pool}
                \STATE $P_e \leftarrow $ \textbf{PoolUpdate}( $\textbf{M}_{e,t}$, $\textbf{M}_{e,t-1}$)

                \IF{\textbf{PLAN\_REQUIRED}} 
                    \STATE \textcolor[HTML]{6F9759}{\# Search a new goal from the maps}
                    \STATE \textbf{GoalSearch}($\textbf{G}_{t}$, $P_e$, $T_t$) $\rightarrow$ $T_g$ 
                    
                    \STATE \textcolor[HTML]{6F9759}{\# Plan a feasible path based on $\textbf{M}_{e,t}$ towards $T_g$}
                    \STATE \textbf{PathPlanning}($\textbf{M}_{e,t}$, $T_t$, $T_g$) $\rightarrow$ $\{T_j\}_{j=t}^g$
    
                    \STATE \textcolor[HTML]{6F9759}{\# Set \textbf{PLAN\_REQUIRED} to False}
                    \STATE \textbf{PLAN\_REQUIRED} $\leftarrow$ False
                \ENDIF

                \STATE \textcolor[HTML]{6F9759}{\# Update pose from planned path}
                \STATE \textbf{Action}  $T_t \leftarrow \{T_j\}_{j=t}^g$

                \STATE \textcolor[HTML]{6F9759}{\# Replanning after reached goal}
                \STATE \textbf{CheckPlanRequired}: update \textbf{PLAN\_REQUIRED}
        \ENDWHILE
        \STATE $t_1 = t $
        \STATE \textcolor[HTML]{979759}{\# POST-REFINEMENT}
        \FOR {$t \leftarrow \text{$t_1$ to $t_1 + T$}$}

                \STATE \textcolor[HTML]{6F9759}{\# Post-refinement based on $\{O\}_{i=0}^{t_1}$}
                \STATE \textbf{Map Update}:  $\textbf{G}_{t} \leftarrow \textbf{G}_{t-1}$
        
        \ENDFOR
  \end{algorithmic}
\label{alg:ActiveGS}
\end{algorithm} 
  

\subsection{Global-Local Keyframe Selection}
\label{sec:method:keyframe_selection}

Following SplaTAM \cite{keetha2024splatam}, we design our Gaussian Mapping backbone as detailed in \cref{sec:method:gaussian_map}.
Instead of optimizing with every frame, SplaTAM selects every fifth frame as a keyframe and updates the map using a subset of keyframes that overlaps with the current frame.
This approach ensures that the map is optimized with keyframes that influence newly added Gaussians, providing sufficient multiview supervision without processing all prior keyframes.

The $k$ selected local keyframes include the current frame, the last keyframe, and $k-2$ keyframes with the highest overlap with the current frame.
Overlap is determined by projecting the current frame's depth map into a point cloud and counting points within each keyframe's frustum.
This local keyframe strategy yields accurate scene representation, especially in recently updated regions.
However, it also causes potential overfitting in local areas.
Gaussians that do not contribute directly to rendering may have their opacity minimized, particularly those within the view frustum but behind the primary surface.

To mitigate this issue, we propose a global-local keyframe selection strategy that balances local supervision with global regularization.
Global keyframes are selected based on the information they provide, determined by two criteria:
(1) Completeness: A keyframe is included if it reveals over 10\% new pixels in the silhouette mask.
(2) Quality: A keyframe is included if its rendering quality falls below a threshold.
Type-1 global keyframes primarily support comprehensive scene reconstruction, while Type-2 keyframes focus on challenging regions.
In conclusion, we select half of the keyframes from local overlapping frames and the other half from global supporting keyframes to maintain an effective balance of local detail and global structural accuracy.

\subsection{Post-Refinement}
\label{sec:post_refinement}
The exploration stage focuses on achieving completeness in both geometric (3D) and photometric (rendering) reconstruction.
However, rendering quality may be suboptimal due to limited optimization of the Gaussian Map.
To address this, we introduce a post-refinement step, leveraging the global keyframes from \cref{sec:method:keyframe_selection} to enhance photometric reconstruction quality.

We note, however, that post-refinement can degrade geometric reconstruction, as Gaussian Map optimizations for rendering involve operations that are unfavorable for preserving geometric detail.
Specifically, optimizing the Gaussian Map for photometric quality may result in Gaussian pruning in low-texture areas and removal of redundant Gaussians, reducing geometric completeness.
Despite these trade-offs, we observe overall improvements in rendering quality with post-refinement.

\subsection{\titlePrefix{} System}
\label{sec:method:active_gs}

After capturing new RGB-D frames, a selection of keyframes is stored in keyframe a database to optimize mapping.
Integrating the mapping module from \cref{sec:method:gaussian_map} and \cref{sec:method:keyframe_selection} with the planning module in \cref{sec:method:render_planning}, we establish a comprehensive Active Gaussian Mapping system, namely \textit{\titlePrefix{}},  detailed in \cref{alg:ActiveGS} and illustrated in \cref{fig:framework}.
This system maintains an up-to-date Gaussian Map in an incremental manner, leveraging the planning module for goal searching and path planning.

\section{Experiments and Results}
\label{sec:exp}

\subsection{Experimental Setup}
\label{sec:exp:exp_setup}
\paragraph{Simulator and Dataset}

Our experiments are conducted in the Habitat simulator \cite{savva2019habitat} and evaluated on two photorealistic datasets: Replica \cite{straub2019replica} and Matterport3D (MP3D) \cite{chang2017matterport3d}.
We use 8 scenes from Replica \cite{sucar2021imap} and 5 scenes from MP3D \cite{yan2023active} for analysis.
Each experiment runs for 2000 steps in Replica and 5000 steps in MP3D, with the extended steps in MP3D reflecting its larger scene sizes and need for thorough exploration.

In these experiments, our system processes posed RGB-D images at a resolution of $680 \times 1200$, with vertical and horizontal fields of view set at $60^\circ$ and $90^\circ$, respectively.
The voxel size for generating the Exploration Map is fixed at 5 cm across all experiments.

Unlike prior active radiance field methods, which often constrain actions to teleporting between discrete locations \cite{pan2022activenerf, ran2022neurar}, moving within hemispheres \cite{zhan2022activermap}, or navigating limited 2D planes \cite{chaplot2020activeneuralslam, yan2023active}, our approach enables 6DoF movement in unrestricted 3D spaces.

\paragraph{Geometric Metrics}
We evaluate geometric reconstruction using three metrics: \textit{Accuracy} (cm), \textit{Completion} (cm), and \textit{Completion ratio} (\%) with a 5 cm threshold.
To calculate these metrics, we evenly sample 3D points from the ground-truth meshes and compare them with a point cloud uniformly extracted from the Gaussian Maps.

\paragraph{Rendering Metrics}
For measuring RGB rendering performance we use PSNR, SSIM and LPIPS. 
For depth rendering performance we use Depth L1 distance. 

\subsection{Evaluation}
\label{sec:exp:benchmark}

\begin{table}[t]
    \centering
    \resizebox{1\columnwidth}{!}{
    \begin{tabular}{l c c c c}
        \hline
         \textbf{Methods }& \textbf{Acc. (cm)} $\downarrow$ & \textbf{Comp. (cm)} $\downarrow$ & \textbf{Comp. Ratio (\%)} $\uparrow$ \\
        \hline
        FBE \cite{yamauchi1997frontier} &
        / & 9.78 & 71.18 \\
        
        UPEN \cite{georgakis2022uncertainty} &
        / & 10.60 & 69.06 \\
        
        OccAnt \cite{ramakrishnan2020occupancy} &
        / & 9.40 & 71.72 \\
        
        \textcolor[HTML]{A24F78}{ANM \cite{yan2023active}} &
        7.80 & 9.11 & 73.15 \\
        
        \textcolor[HTML]{98C281}{\textbf{NARUTO}\cite{feng2024naruto}} &
        6.31 & 3.00 & 90.18 \\

        \textcolor[HTML]{ACD0FD}
        {\textbf{Ours}} &
        \textbf{1.66} & \textbf{2.30} & \textbf{95.32} \\
        \hline
    \end{tabular}
    }
    \caption{ \textbf{MP3D Results.}
    Our method shows superior performance with better reconstruction quality and completeness.
    }
    \label{tab:mp3d}
    \vspace{-10pt}
\end{table}
\begin{figure*}[th!]
        \centering
		\includegraphics[width=1.0\textwidth]    {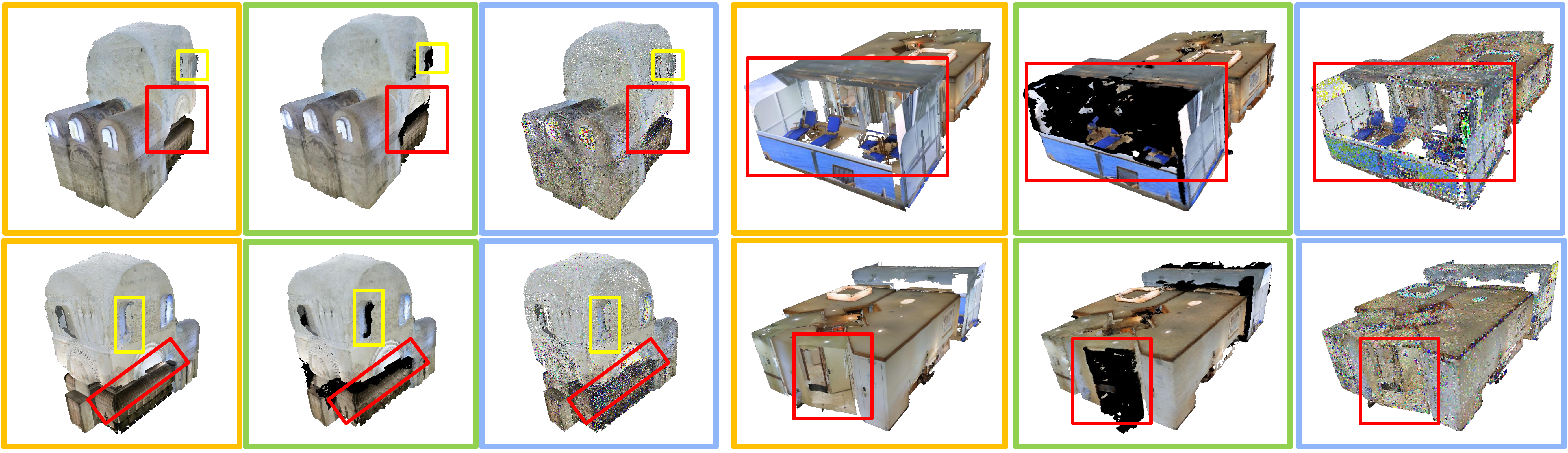}
		\caption{
            \textbf{3D Reconstruction Results on MP3D.} 
            Shown are two scenes (Left: pLe4; Right: HxpK) with results distinguished by border colors: [\textcolor[HTML]{F4B070}{Ground Truth}, \textcolor[HTML]{98C281}{NARUTO} \cite{feng2024naruto}, \textcolor[HTML]{ACD0FD}{Ours}]. 
            For NARUTO, black regions indicate neural mapping extrapolation, sometimes with inaccuracies. 
            For our method, colored point clouds are extracted from the Gaussian Map for visualization; 
            note that noisy points here do not reflect actual rendering quality. 
            Unlike neural maps, the explicit Gaussian Map representation avoids inaccurate extrapolation artifacts.
            }
            
		\label{fig:mp3d_result}
		\vspace{-10pt}
\end{figure*}

To the best of our knowledge, this work is among the \textit{first} to address active mapping using 3D Gaussian Splatting. 
Concurrently, recent technical reports \cite{jiang2024agslam, li2024activesplat} explore related concepts, highlighting the growing interest in this area.

As discussed in \cref{sec:post_refinement}, the Refined Model enhances rendering performance but reduces geometric completeness by removing redundant Gaussians.
To accurately assess both aspects, we evaluate geometric performance using the Exploration Model and rendering performance using the Refinement Model. 
The choice of model depends on the specific application requirements, balancing the need for detailed geometry versus high-quality rendering.

\begin{figure*}[t!]
        \centering
		\includegraphics[width=1.\textwidth]{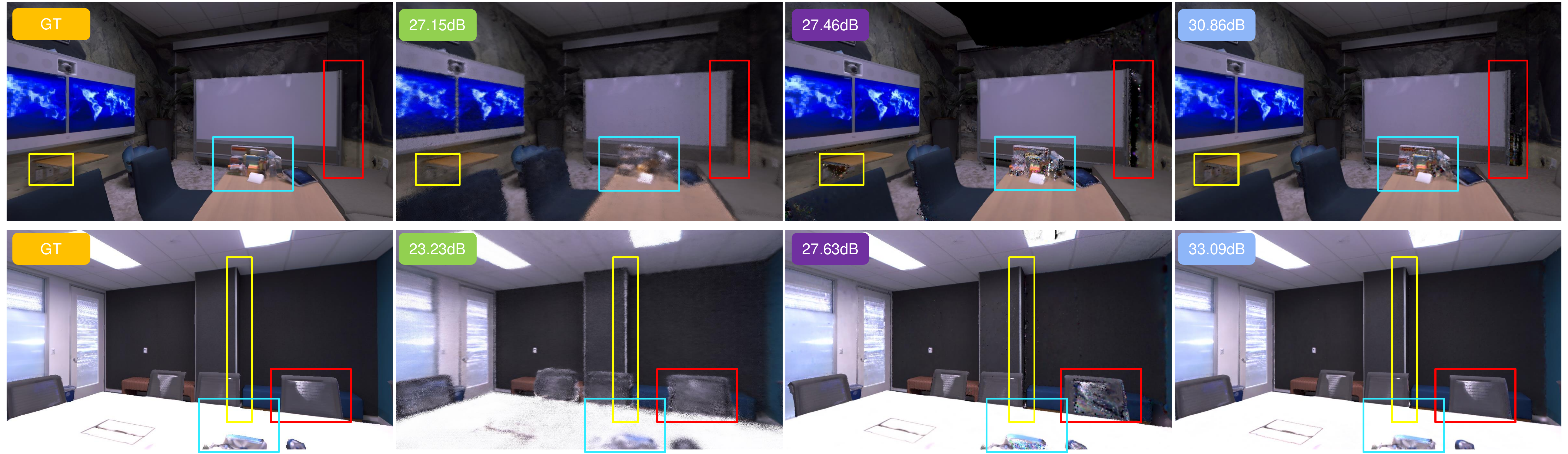}
		\caption{
            \textbf{Rendering Results on Replica.} 
            Two scenes (office0, office4) are shown in the first and second rows, respectively.
            The results represent 
            [\textcolor[HTML]{F4B070}{Ground Truth},
            \textcolor[HTML]{98C281}{NARUTO},
            \textcolor[HTML]{703691}{SplaTAM w/o tracking},
            \textcolor[HTML]{ACD0FD}{Ours}].
            Our renders demonstrate higher fidelity across most regions.
            }
            \label{fig:replica_result}
		\vspace{-10pt}
\end{figure*}

\paragraph{Geometric Evaluation}
\cref{tab:mp3d} presents a quantitative comparison of our system against prior studies on MP3D.
Our approach consistently outperforms previous methods across all evaluation metrics.
The accuracy metric reflects the geometric precision of the Gaussian Map, which, given our initialization with observed depths, achieves high accuracy as expected.
Additionally, both the Completion and Completion Ratio metrics, which measure the 3D space coverage through active exploration, show that our method attains exceptional completeness.
This achievement stems from our rendering-based planning, which efficiently identifies the most informative viewpoints, combined with the agent's unrestricted movement capabilities.
We also present qualitative comparisons in \cref{fig:mp3d_result}. 

\begin{table}[t!]
\centering
\resizebox{1\columnwidth}{!}{
    \begin{tabular}{lcccccccccc}
    \toprule
    \textbf{Methods} & \textbf{Metrics} & \textbf{Avg.} & \textbf{Of0} & \textbf{Of1} & \textbf{Of2} & \textbf{Of3} & \textbf{Of4} & \textbf{R0} & \textbf{R1} & \textbf{R2}  \\
    \midrule
    \multirow{4}{*}{\textcolor[HTML]{703691}{SplaTAM} \cite{keetha2024splatam}} 

    & PSNR $\uparrow$ & 29.08 & 30.15 & 33.60 & 26.40 & 23.97 & 29.87 & 31.13 & 29.74 & 27.79 \\
    & SSIM $\uparrow$ & 0.95 & 0.96 & 0.96 & 0.94 & 0.90 & 0.96 & 0.97 & 0.96 & 0.96 \\
    & LPIPS $\downarrow$ & 0.14 & 0.13 & 0.16 & 0.16 & 0.21 & 0.13 & 0.09 & 0.12 & 0.11 \\
    & L1-D $\downarrow$ & 1.38 & 1.34 & 0.89 & 1.75 & 3.28 & 1.26 & 0.64 & 0.84 & 1.02 \\
    \midrule

    \multirow{4}{*}{\textcolor[HTML]{98C281}{NARUTO} \cite{feng2024naruto}} 
    & 
    PSNR $\uparrow$ & 26.01 & 28.88 & 33.27 & 24.26 & 25.32 & 22.75 & 24.70 & 26.17 & 22.77 \\
    & SSIM $\uparrow$ & 0.89 & 0.93 & 0.96 & 0.89 & 0.91 & 0.88 & 0.87 & 0.90 & 0.82 \\
    & LPIPS $\downarrow$ & 0.41 & 0.40 & 0.27 & 0.38 & 0.38 & 0.44 & 0.49 & 0.43 & 0.52 \\
    & L1-D $\downarrow$ & 9.54 & 4.51 & 1.35 & 5.16 & 6.28 & 7.76 & 6.07 & 4.08 & 41.14 \\
    \midrule

    \multirow{4}{*}{\textcolor[HTML]{ACD0FD}{Ours}} 

    & PSNR $\uparrow$ 
    & \textbf{32.02} & 33.70 & 35.78 & 29.98 & 32.17 & 33.59 & 29.64 & 30.25 & 31.03 \\
    & SSIM $\uparrow$ 
        & \textbf{0.97} & 0.97 & 0.97 & 0.97 & 0.98 & 0.98 & 0.96 & 0.96 & 0.97 \\
    & LPIPS $\downarrow$ 
        & \textbf{0.11} & 0.10 & 0.13 & 0.10 & 0.08 & 0.08 & 0.14 & 0.12 & 0.10 \\
    & L1-D $\downarrow$ 
        & \textbf{1.12} & 1.38 & 0.86 & 1.70 & 0.99 & 0.87 & 0.99 & 1.01 & 1.21 \\
    \midrule

    \end{tabular}
}
\caption{
\textbf{Novel View Rendering Performance on Replica \cite{straub2019replica}.}
}
\label{tab:replica_nvs}
\end{table} 

\paragraph{Rendering Evaluation}
We evaluate rendering performance on the high-fidelity Replica Dataset, as shown in \cref{tab:replica_nvs}, using novel view trajectories for Novel View Rendering.
Two baselines are used for comparison.

The first baseline is the passive mapping method, SplaTAM \cite{keetha2024splatam}.
For a fair comparison, we disable its tracking thread and simulate handheld scanning with a manually defined capture trajectory.
Since manual scanning doesn’t cover the entire scene, we exclude uncovered regions from the evaluation, focusing only on captured areas.

The second baseline is NARUTO \cite{feng2024naruto}, a state-of-the-art active mapping method that uses a neural radiance field.
As NARUTO actively maps the environment, we evaluate all pixels in the novel views to assess its coverage fully.

Our proposed method consistently outperforms both baselines by a significant margin, benefiting from \textit{actively} building a \textit{Gaussian Map} representation capable of high-fidelity rendering, as evidenced in the quantitative results and qualitative examples in \cref{fig:replica_result}.

\subsection{Ablation Studies} 
\label{sec:exp:ablation}

\begin{figure}[t!]
        \centering
  	\includegraphics[width=1.0\columnwidth]{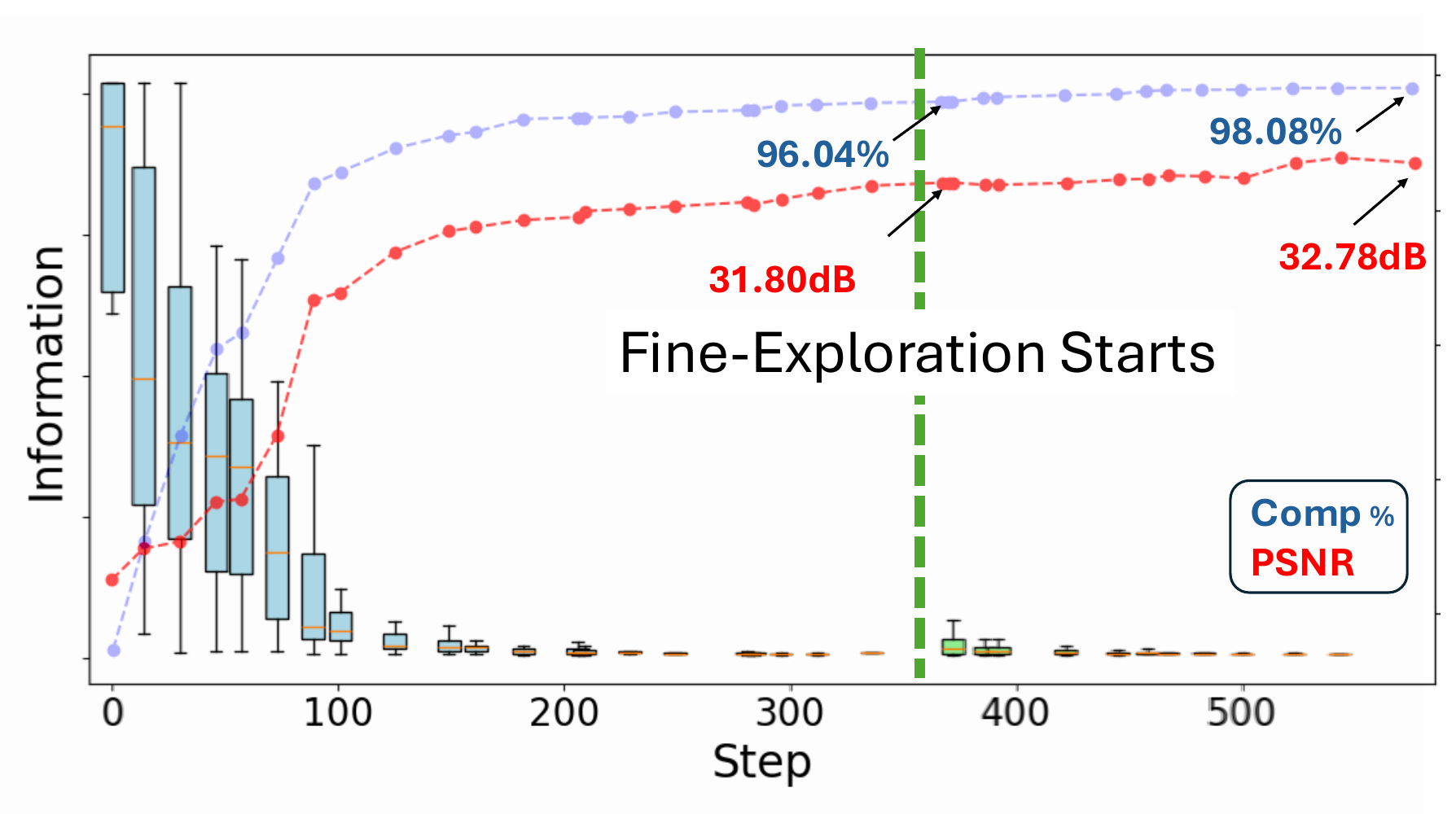}
    \vspace{-20pt}
		\caption{
            \textbf{Reconstruction Progress in Replica-office0}. 
            The distribution of rendering-based information ($N_{S_i}$) of candidate viewpoints is shown alongside reconstruction performance metrics. 
            As information decreases, completeness and rendering performance improve, with
            Fine-Exploration yielding an additional boost.
            }
		\label{fig:completeness_progress}
		\vspace{-10pt}
\end{figure}

The Replica dataset features photorealistic 3D indoor scenes, represented by dense meshes with higher completeness than MP3D scenes.
Given this level of detail, we primarily conduct our ablation studies on Replica to obtain more representative and robust results.
In the following experiments, we use \textbf{Full} as the reference system, with ablation studies performed with respect to this configuration.
The results are shown in \cref{tab:ablation}.

\paragraph{Coarse-to-Fine Exploration}
We first investigate the effectiveness of the coarse-to-fine exploration strategy.
While Coarse Exploration alone achieves substantial scene completeness, Fine Exploration—denoted as ``w/o Refine"—further enhances both geometric and photometric reconstruction.
\cref{fig:completeness_progress} illustrates the progression of exploration completeness.
As shown, our method quickly reaches high completeness in the Coarse Exploration stage and then refines missing details in the Fine Exploration stage.

\paragraph{Post-Refinement}
As discussed in \cref{sec:post_refinement}, the post-refinement step in our full method improves rendering quality by further optimizing the Gaussian Map based on a rendering-oriented losses.
However, this step may lead to Gaussian pruning, reducing the completeness of the map.
When Post-Refinement is omitted (``w/o Refine"), the geometric reconstruction metrics improve due to the retention of redundant Gaussians, which are beneficial for geometry.
Nonetheless, the post-refinement step significantly enhances rendering performance.

\paragraph{Global-Local Keyframe Selection}
In the experiment ``w/o Global-KF," we assess the impact of excluding Global Keyframes.
Without the regularization from Global Keyframes, 3D reconstruction performance declines, underscoring the importance of supervising by local and global keyframes to optimize the Gaussian Map effectively.

\begin{table}[t]
    \centering
    \resizebox{1\columnwidth}{!}{
    \begin{tabular}{l c c c c c}
        \hline
         \textbf{Exp.} & \textbf{Comp. (cm)} $\downarrow$ & \textbf{Comp. Ratio (\%)} $\uparrow$ & \textbf{PSNR (dB)} $\uparrow$ & \textbf{L1-D (cm)}$\downarrow$\\
        \hline

        {Coarse Exploration} &
        1.77 & 94.53 & 29.77 & 1.80 \\

        {w/o Global-KF} &
        2.19 & 94.87 & 30.73 & 1.23  \\

        {w/o Refine.} &
        \textbf{1.56} & \textbf{96.50} &  30.67 & 1.42   \\

        {Full} &
        1.80 & 95.45 & \textbf{32.02} & \textbf{1.12}  \\
        \hline
        
    \end{tabular}
    }
    \caption{ \textbf{Ablation Studies on Replica.}
    }
    \label{tab:ablation}
    \vspace{-10pt}
\end{table}
\section{Discussion}
\label{sec:conclusion}

In this paper, we introduced \textit{\titlePrefix{}}, an active mapping system that leverages 3D Gaussian Splatting (3DGS) for efficient, high-fidelity real-time scene reconstruction and exploration. By harnessing the fast rendering capabilities of 3DGS, our approach overcomes limitations of traditional NeRF-based methods, enabling unrestricted 6DoF movement and enhancing both geometric completeness and photometric fidelity.

Key features of \textit{\titlePrefix{}} include a rendering-based information gain module for optimal viewpoint selection and a balanced framework incorporating coarse-to-fine exploration, post-refinement, and global-local keyframe selection. Together, these components enable autonomous, high-quality exploration and reconstruction in complex environments, achieving state-of-the-art accuracy and fidelity. Extensive evaluations on challenging datasets like Replica and MP3D confirm that \textit{\titlePrefix{}} outperforms existing methods in both geometric accuracy, reconstruction completeness, and rendering quality, highlighting 3DGS as a valuable tool for real-time active vision applications

While \textit{\titlePrefix{}} demonstrates strong performance, several directions for future work could enhance its applicability. 
First, real-world scenarios require a robust planning and localization module to address assumptions of known localization and perfect action execution.
Second, to increase system versatility, future iterations should account for real-world motion constraints that affect the agent's movement. 
Additionally, the current rendering-based information gain approach may overlook certain regions, such as double-sided objects. For example, after reconstructing one side of an object, the system may not detect the need to explore the back side if it’s not visible in the initial view, leading to incomplete reconstructions. Future work will aim to integrate additional semantic and surface cues to better understand scene complexity and enhance exploration guidance. 

\clearpage
\setcounter{page}{1}
\maketitlesupplementary


\section{Overview}
\label{sec:rationale}

In this supplementary material, we provide a detailed outline structured as follows:  
\cref{supp:sec:implementation} delves into additional implementation specifics of \titlePrefix{}.  
\cref{supp:sec:runtime} examines the computation costs associated with each module.  
Complementing the results in \cref{sec:exp}, \cref{supp:sec:more_dataset_results} extends our analysis for MP3D and Replica, including per-scene reconstruction and novel view rendering results, both quantitative and qualitative.  
We also include trajectory visualizations to illustrate the planned trajectories.  
Lastly, we present analysis and visualizations of common failure cases observed in our method.

\section{Implementation Details}
\label{supp:sec:implementation}

\subsection{Hardware Requirements}
We conduct our experiments on a desktop PC equipped with a 2.2GHz Intel Xeon E5-2698 CPU and an NVIDIA V100 GPU.  
Memory consumption varies depending on the scene size.  
For reference, in an $80 \, \text{m}^3$ scene, the GPU memory and RAM usage are as follows across different stages:

\begin{itemize}
    \item \textbf{Coarse-Exploration}: GPU memory consumption is approximately 5.3GB, and RAM usage is 1GB.
    \item \textbf{Fine-Exploration}: GPU memory increases to 6.2GB due to the addition of more Gaussians.
    \item \textbf{Post-Refinement}: GPU memory reaches 6.5GB, and RAM usage increases to 1.8GB as more optimization iterations are performed.
\end{itemize}

\subsection{Coarse-to-fine Exploration Details} \label{supp:sec:exploration_details}
\paragraph{Avoiding candidate sampling near surfaces. }
While building the Exploration Candidate Pool, we intentionally avoid sampling candidates from free spaces close to surfaces for two key reasons:
(1) Rendering Limitations: The Gaussian Map renderer ignores 3D Gaussians near the camera. When candidates are close to surfaces (e.g., walls), the most informative viewpoints tend to face the surface directly. This results in low-visibility masks due to the renderer ignoring nearby Gaussians, leading to incorrect candidate selection.
(2) Collision Prevention: To simulate collision avoidance between the agent/camera and the surface, we maintain a buffer of free space between them.

\subsection{Post-Refinement Details}
During the exploration stages, each Gaussian Map update step uses only 15 optimization iterations, and a quarter of the image resolution is employed for densification.
In the post-refinement stage, the number of optimization iterations increases from 15 to 60, and the full image resolution is utilized for densification.

\section{Runtime Analysis} \label{supp:sec:runtime}
\begin{table*}[t!]
\centering
\resizebox{0.8\textwidth}{!}{
\begin{tabular}{l l c c}
\hline
\textbf{Stage} & \textbf{Module}        & \textbf{Per Iteration (s)} & \textbf{Per Stage (s)} \\ \hline \hline

\textbf{All} & HabitatSim & 0.244 & 0.244 \\
\hline \hline

\multirow{3}{*}{\textbf{Coarse-Exploration}} 
               & Gaussian Mapping       & 0.687                &  0.181 \\ 
               \cline{2-4} 
               & Planning - Global       & 1.126                & \multirow{2}{*}{0.071}  \\ 
               & Planning - Local        & 0.018               &  \\ \hline \hline
               
\multirow{3}{*}{\textbf{Fine-Exploration}}  
               & Gaussian Mapping       & 0.759                & 0.184 \\ 
               \cline{2-4} 
               & Planning - Global       & 5.411                & \multirow{2}{*}{0.283}  \\ 
               & Planning - Local        & 0.023               &  \\ \hline \hline
\textbf{Post-Refinement} & Gaussian Mapping & 2.580                  & 2.580                       \\ \hline
\end{tabular}
}
\caption{\textbf{Runtime Analysis of \titlePrefix{} Modules on Replica-room0.}}
\label{tab:runtime_analysis}
\end{table*}
In this section, we present a runtime analysis of the three major modules in \titlePrefix{}:
\begin{itemize}
    \item \textbf{Data Generation}: A simulator generates RGB-D data.
    \item \textbf{Gaussian Mapping Module}: Updates the Gaussian Map.
    \item \textbf{Rendering-Based Planning Module}: Includes a global planner for goal searching and a local planner for path planning.
\end{itemize}

For data generation, HabitatSim is required to generate $680 \times 1200$ RGB-D data per iteration.  
The detailed runtime for each stage is presented in \cref{tab:runtime_analysis}.  
We compute the statistics using average.  

Two types of statistics are reported in the table:  
\begin{itemize}
    \item \textbf{Per-iteration timing}: The runtime for each individual module when it is activated.  
    \item \textbf{Overall timing per stage}: The average runtime considering the infrequent activation of certain modules.  
\end{itemize}

The \textbf{Gaussian Mapping Module} is active only during keyframe steps (\ie every five frames) in the exploration stage, reducing its average runtime.  
The \textbf{Planning Module} is triggered on demand, contributing minimal runtime overhead.  
In contrast, \textbf{Post-Refinement} involves more intensive computation with additional iterations.

\section{Additional Experimental Results}
\label{supp:sec:more_dataset_results}




\subsection{Per-scene Results on Replica}
\begin{table*}[htbp]
    \centering
    \resizebox{1\textwidth}{!}{
    \begin{tabular}{l l*{8}{c}|c}
        \hline
        \textbf{Methods} & \textbf{Metrics} & 
        office0 & office1 & office2 & office3 & office4 & room0 & room1 & room2 & 
        Avg. \\
        \hline
        \hline

        \multirow{3}{*}{
        \textcolor[HTML]{98C281}{\textbf{NARUTO}}\cite{feng2024naruto}
        } 
        & Acc. (cm) $\downarrow$ & 
        1.30 & 1.03 & 2.25 & 2.29 & 1.75 & 1.56 & 1.25 & 1.47 & 1.61 \\
        & Comp. (cm) $\downarrow$ & 
        1.39 & 1.53 & 1.69 & 2.27 & 1.79 & 1.68 & 1.43 & 1.48 & 1.66 \\
        & Comp. Ratio $\uparrow$ & 
        98.17 & 95.26 & 97.54 & 93.91 & 97.93 & 98.28 & 98.04 & 98.47 & \textbf{97.20} \\
        \hline

        \multirow{3}{*}{
        \textcolor[HTML]{ACD0FD}{\textbf{Ours}}
        } 
        & Acc. (cm) $\downarrow$ & 
        1.35 & 1.07 & 1.06 & 1.03 & 0.84 & 1.22 & 1.40 & 1.31 & \textbf{1.16} \\
        & Comp. (cm) $\downarrow$ & 
        1.93 & 1.26 & 1.31 & 1.13 & 1.02 & 1.89 & 2.06 & 1.89 & \textbf{1.56}\\
        & Comp. Ratio (\%) $\uparrow$ & 
        96.29 & 98.14 & 97.91 & 98.12 & 98.18 & 94.33 & 93.93 & 95.15 & 96.50 \\
        \hline

        \hline

    \end{tabular}
    }
    \caption{\textbf{3D Reconstruction Results on Replica\cite{straub2019replica}.}}
    \label{tab:replica_full_3d}
\end{table*}

\begin{figure*}[th!]
        \centering
		\includegraphics[width=1\textwidth]    {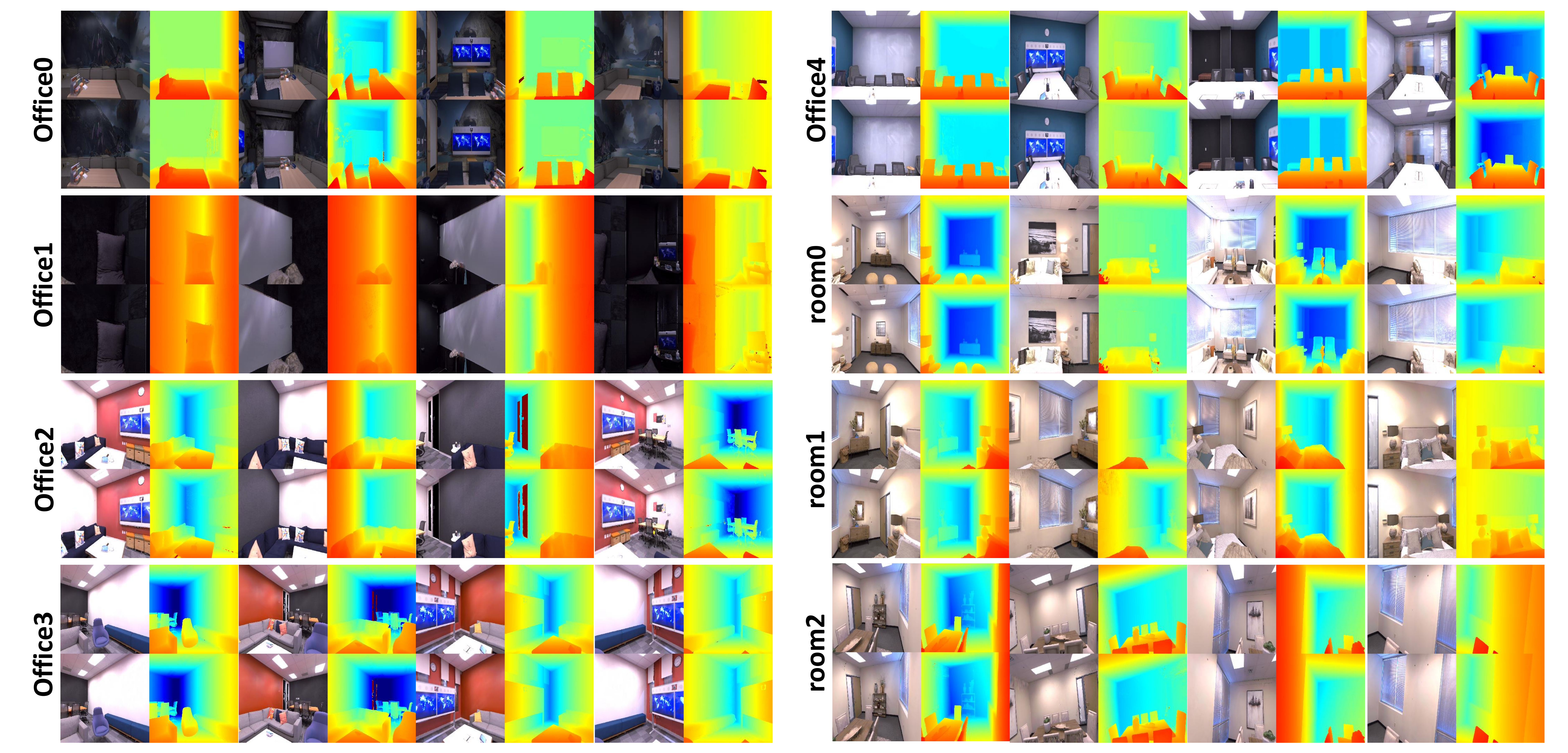}
		\caption{
\textbf{Novel View Rendering Results on Replica.} 
This figure provides a side-by-side comparison of rendering results on the Replica dataset.
Upper rows show the ground truth RGB-D images, while bottom rows present the rendered RGB-D images.
            }
		\label{fig:replica_result_full}
\end{figure*}

\paragraph{3D Reconstruction}
In \cref{tab:replica_full_3d}, we present the 3D reconstruction results on the Replica dataset \cite{straub2019replica}, comparing our method with the state-of-the-art active mapping approach, NARUTO~\cite{feng2024naruto}.
Our method demonstrates comparable reconstruction performance to NARUTO, achieving higher accuracy and completeness on average while maintaining a similar completeness ratio.
The slightly lower completeness ratio in our results may be attributed to the extrapolation capability inherent in NARUTO's neural radiance field representation.

\paragraph{Qualitative Results}

In \cref{fig:replica_result_full}, we present a qualitative evaluation of our method's rendering performance against the ground truth across various scenes in the Replica dataset.
The evaluation involves a novel view trajectory with a $360^\circ$ circular movement (an outside-in trajectory looking at a specific point in the scene).
To provide a comprehensive visualization, we present the rendering results from four directions: front, left, right, and back.
Our method demonstrates high fidelity in both RGB and depth rendering.

\subsection{Per-scene Results on MP3D}
\begin{table*}[t]
    \centering
    \resizebox{0.8\textwidth}{!}{
    \begin{tabular}{l l  *{5}{c} c}
        \hline
        \textbf{Methods} & \textbf{Metrics} & 
        \textbf{Gdvg} & \textbf{gZ6f} & \textbf{HxpK} & \textbf{pLe4} & \textbf{YmJk} & 
        \textbf{Avg.} \\
        \hline
        \multirow{4}{*}{\textcolor[HTML]{A24F78}{ANM} \cite{yan2023active}} 
        & Acc. (cm) $\downarrow$ & 
        5.09 & 4.15 & 15.60 & 5.56 & 8.61 & 7.80 \\
        & Comp. (cm) $\downarrow$ & 
        5.69 & 7.43 & 15.96 & 8.03 & 8.46 & 9.11   \\
        & Comp. Ratio $\uparrow$ & 
        80.99 & 80.68 & 48.34 & 76.41 & 79.35 & 73.15   \\
        \hline

        \multirow{3}{*}{\textcolor[HTML]{98C281}{NARUTO}\cite{feng2024naruto}} 
        & Acc. (cm) $\downarrow$ & 
        3.78 & 3.36 & 9.24 & 5.15 & 10.04 & 6.31 \\
        & Comp. (cm) $\downarrow$ & 
        2.91 & 2.31 & 2.67 & 3.24 & 3.86 & 3.00 \\
        & Comp. Ratio $\uparrow$ & 
        91.15 & 95.63 & 91.62 & 87.76 & 84.74 & 90.18 \\
        \hline

        \multirow{3}{*}{\textcolor[HTML]{ACD0FD}{Ours}} 
        & Acc. (cm) $\downarrow$ & 
        1.28 & 1.55 & 1.46 & 2.13 & 1.87 & \textbf{1.66} \\
        & Comp. (cm) $\downarrow$ & 
        2.62 & 1.74 & 1.68 & 2.63 & 2.83 & \textbf{2.30} \\
        & Comp. Ratio $\uparrow$ & 
        97.62 & 97.74 & 97.29 & 93.96 & 89.87 & \textbf{95.32} \\
        \hline
    \end{tabular}
    }
    \caption{
    \textbf{3D Reconstruction Results on Matterport3D \cite{chang2017matterport3d} dataset}. 
    Our method achieves consistently better reconstruction than prior methods.
    }
    \label{tab:mp3d_full}
\end{table*}

\begin{table*}[t!]
\centering
\resizebox{0.8 \textwidth}{!}{
    \begin{tabular}{llcccccc}
    \toprule
    \textbf{Methods} & \textbf{Metrics} & \textbf{Gdvg} & \textbf{gZ6f} & \textbf{HxpK} & \textbf{pLe4} & \textbf{YmJk} & \textbf{Avg.}   \\
    \midrule
    
    \multirow{4}{*}{\textcolor[HTML]{98C281}{NARUTO} \cite{feng2024naruto}} 
    & PSNR $\uparrow$ 
        & 22.40 & 22.65 & 14.74 & 21.63 & 21.20 & 20.52 \\
    & SSIM $\uparrow$ 
        & 0.81 & 0.85 & 0.45 & 0.74 & 0.75 & 0.72 \\
    & LPIPS $\downarrow$ 
        & 0.50 & 0.48 & 0.76 & 0.62 & 0.56 & 0.58 \\
    & L1-D $\downarrow$ 
        & 5.11 & 6.00 & 9.82 & 9.91 & 8.94 & 7.95 \\
    \midrule

    \multirow{4}{*}{\textcolor[HTML]{ACD0FD}{Ours}} 
    & PSNR $\uparrow$ 
    & 25.51 & 26.87 & 22.48 & 26.75 & 22.20 &\textbf{ 24.76} \\
    & SSIM $\uparrow$ 
    & 0.92 & 0.96 & 0.82 & 0.94 & 0.84 & \textbf{0.90} \\
    & LPIPS $\downarrow$ 
    & 0.20 & 0.15 & 0.34 & 0.19 & 0.37 & \textbf{0.25} \\
    & L1-D $\downarrow$ 
    & 1.92 & 1.54 & 6.42 & 3.59 & 10.67 & \textbf{4.83} \\
    \midrule

    \end{tabular}
}
\caption{\textbf{Novel View Rendering results on Matterport3D\cite{chang2017matterport3d} dataset.} Our method shows consistently better rendering result than NARUTO.}
\label{tab:mp3d_full_nvs}
\end{table*}
\paragraph{3D Reconstruction}
In \cref{tab:mp3d_full}, we present a comparative analysis of our method against the state-of-the-art approaches, Active Neural Mapping (ANM)~\cite{yan2023active} and NARUTO~\cite{feng2024naruto}.  
The results clearly demonstrate that our method outperforms both ANM and NARUTO across all evaluated metrics.  
Notably, our approach achieves significant improvements in reconstruction quality and completeness, surpassing the benchmarks established by prior methods, thanks to the explicit point cloud representation (extracted from Gaussian Map).
This consistent performance advantage highlights the effectiveness of our method in tackling challenging reconstruction scenarios.

\paragraph{Novel View Rendering}
We further evaluate the novel view rendering performance on MP3D, as shown in \cref{tab:mp3d_full_nvs}.  
For this evaluation, we generate a circular novel view trajectory in each selected scene and compare the performance of NARUTO~\cite{feng2024naruto} and our proposed method.  
Leveraging the Gaussian Mapping representation, our method consistently demonstrates superior rendering performance compared to NARUTO, which relies on a neural radiance field representation.

\paragraph{Qualitative Results}
\begin{figure*}[th!]
        \centering
		\includegraphics[width=0.65\textwidth]    {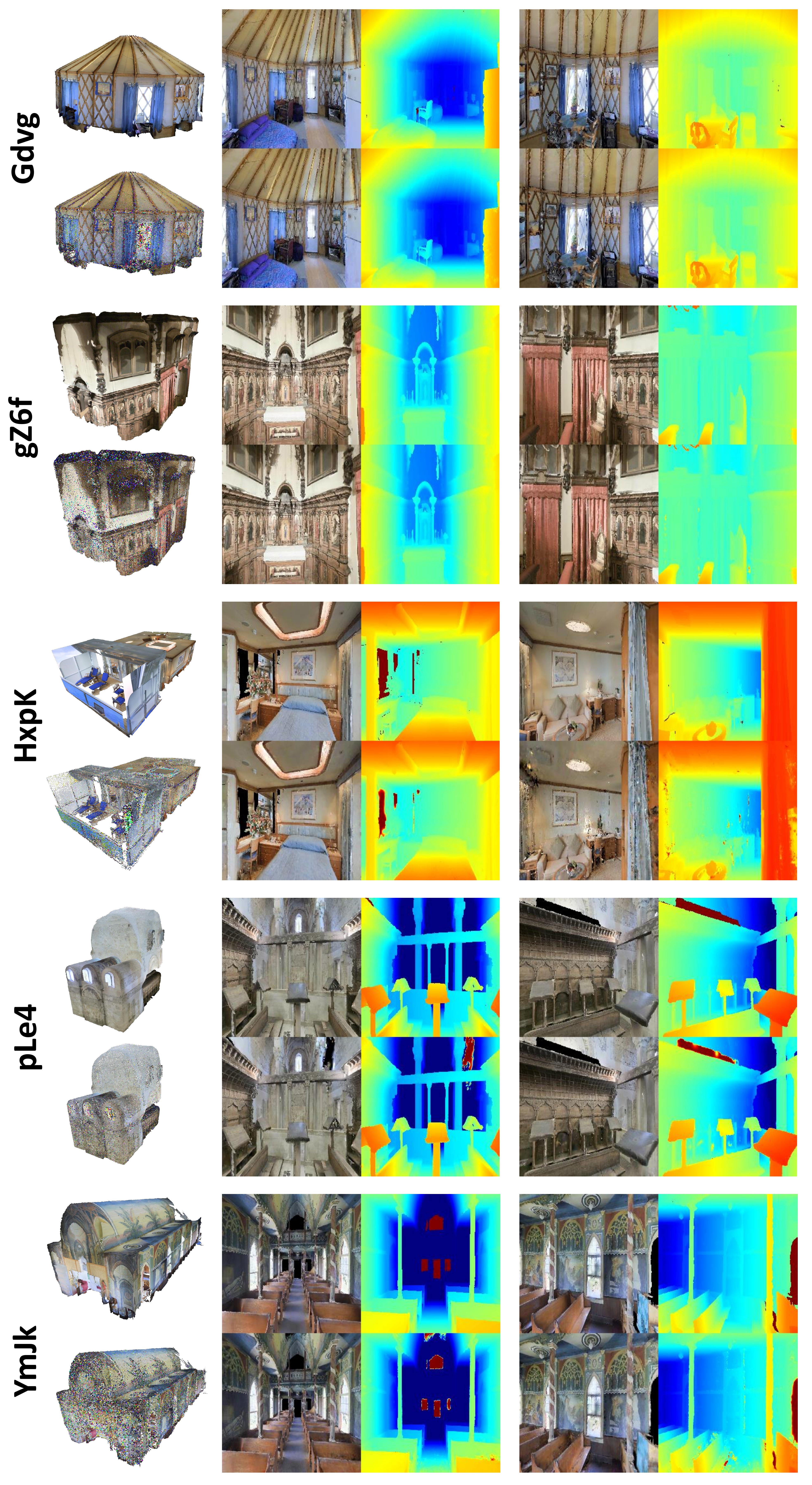}
		\caption{
\textbf{MP3D Reconstruction and Rendering Results.} 
This figure provides a side-by-side comparison of reconstruction results on the MP3D dataset.
Odd-numbered rows show the ground truth meshes and RGB-D images, while even-numbered rows present the 3D point cloud extracted from the Gaussian Map and the rendered RGB-D images.
The noisy points represent low-weight Gaussians and do not affect actual rendering quality.
Our results demonstrate high-quality and complete reconstructions, closely aligning with the ground truths in both geometric and photometric aspects, highlighting the effectiveness of our method in accurately reconstructing complex spatial geometries.
            }
		\label{fig:mp3d_result_full}
\end{figure*}

In \cref{fig:mp3d_result_full}, we present a qualitative evaluation of our method compared to the ground truth for various scenes in the Matterport3D dataset.
The odd-numbered rows display the ground truth data, while the even-numbered rows showcase our method's geometric and photometric reconstructions.
Each scene is labeled with a unique code (\eg, ``Gdvg", ``gZ6f") on the left.
The first column highlights the exterior reconstruction; the second and fourth columns present RGB rendering results of the interior space; and the third and fifth columns show the corresponding depth rendering results.
This result provides a comprehensive visual comparison, effectively illustrating the geometric and photometric performance of our method across the scenes.

\subsection{Trajectory Results}
\begin{figure*}[th!]
        \centering
		\includegraphics[width=1\textwidth]    {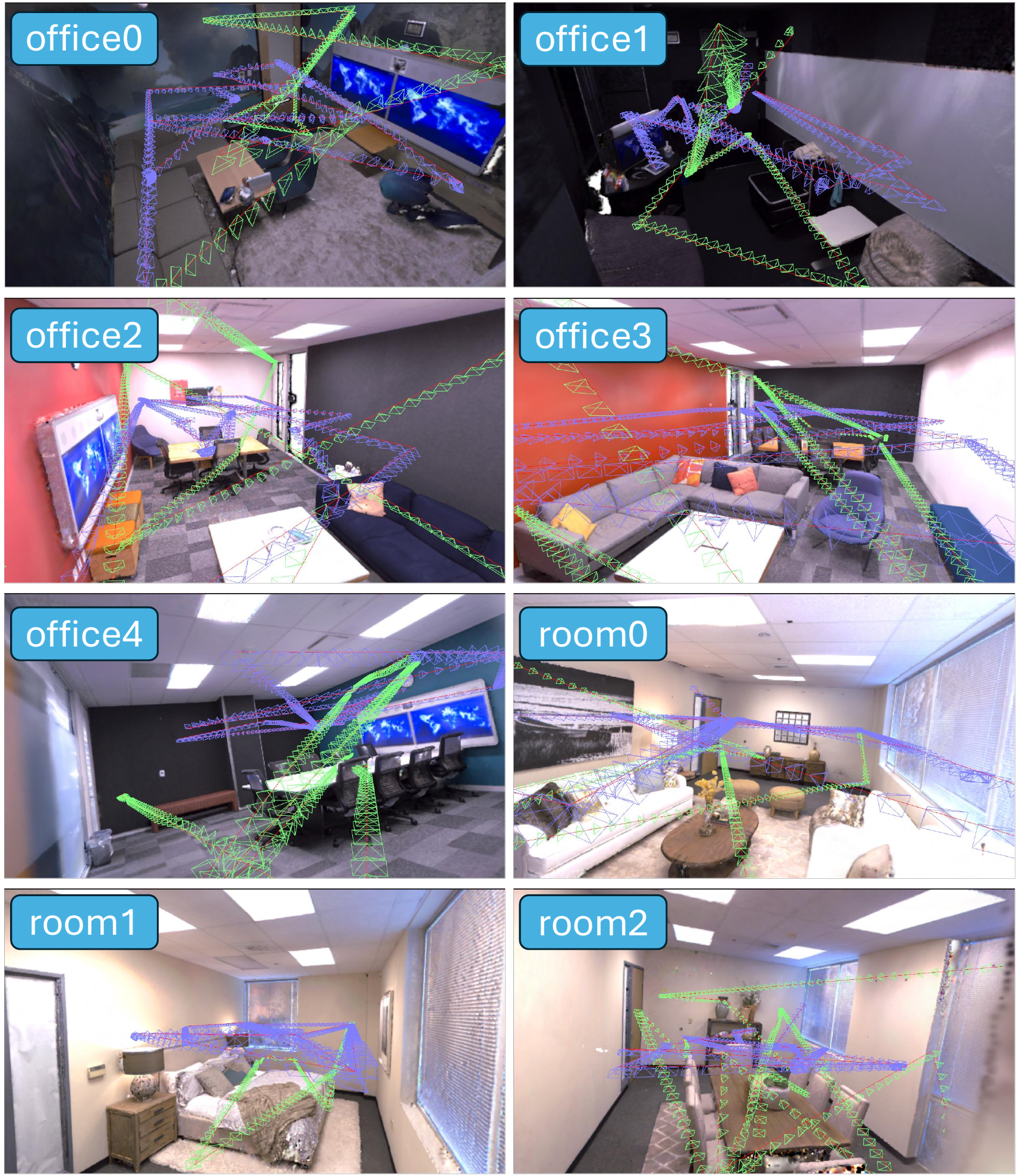}
		\caption{
The exploration trajectories for selected scenes in Replica are visualized, showing only keyframe cameras.
Camera colors indicate the transition from \textcolor[HTML]{8183FB}{coarse} to \textcolor[HTML]{85FC86}{fine} exploration stages.
            }
		\label{fig:traj_replica}
\end{figure*}
We visualize the planned trajectories in Replica scenes, as shown in \cref{fig:traj_replica}.
During \textcolor[HTML]{8183FB}{Coarse-Exploration}, the planned camera movement is restricted to a 2D plane, as candidate sampling is limited to a single height level.
In \textcolor[HTML]{85FC86}{Fine-Exploration}, two height levels are defined, allowing the camera to search for candidates across these levels.

\subsection{Failure Cases}
\begin{figure*}[th!]
        \centering
		\includegraphics[width=1\textwidth]    {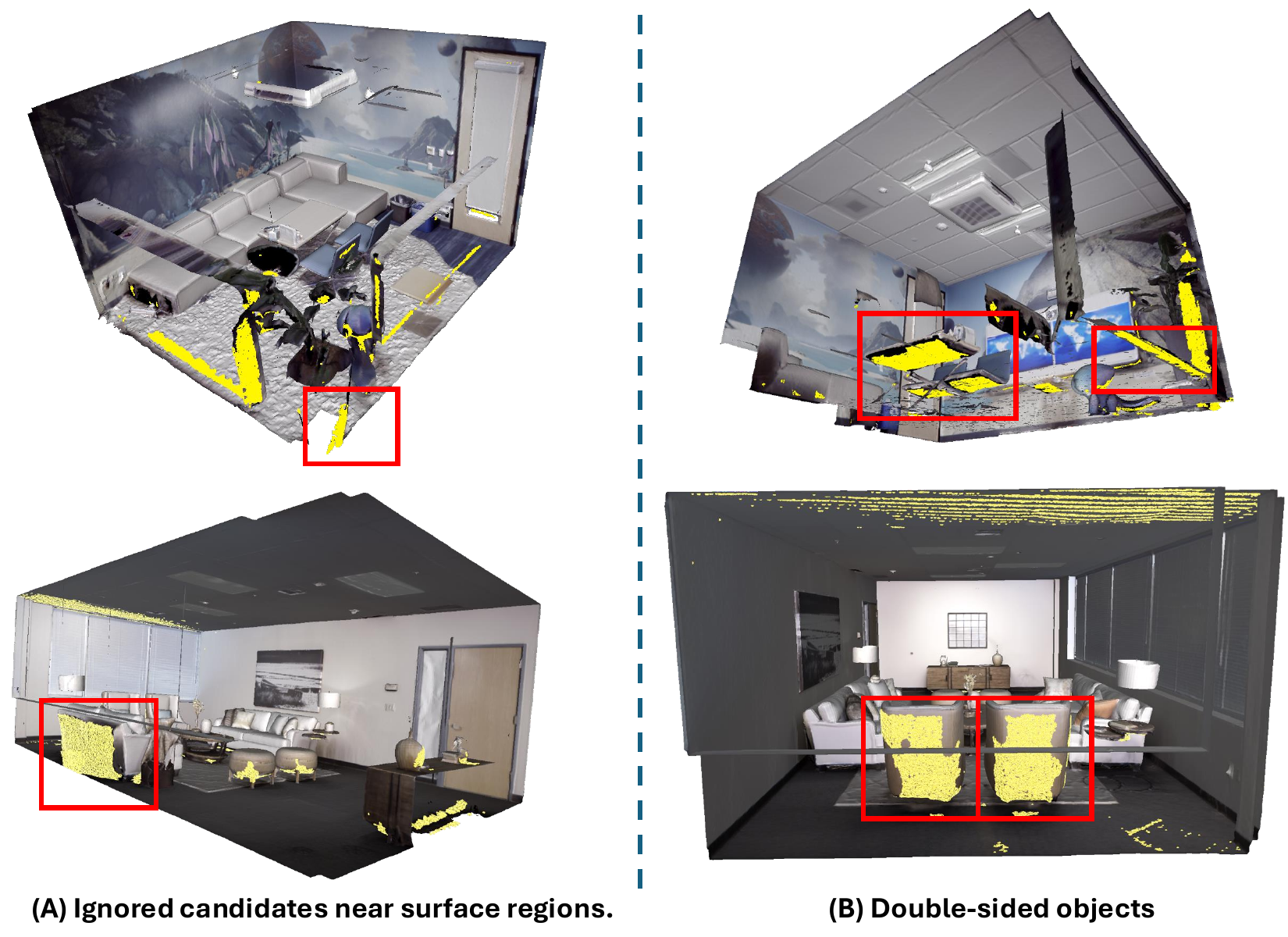}
		\caption{
\textbf{Reconstruction Failure Cases.}
Our method encounters two common failure scenarios:
(A) Insufficient observations due to ignored candidates near surface regions.
(B) Unreconstructed back sides of double-sided objects, as these regions are not captured by the rendering-based information gain.
The yellow-highlighted areas indicate regions with either no reconstruction or low completeness.
            }
		\label{fig:failure_cases}
\end{figure*}

Although our method demonstrates high-quality reconstruction and high-fidelity rendering, some failure cases are observed, as illustrated in \cref{fig:failure_cases}.

\paragraph{Ignored Candidates Near Surface Regions.}
As discussed in \cref{supp:sec:exploration_details}, we avoid sampling candidates from regions near surfaces.  
Consequently, regions that can only be observed from such candidates remain unobserved, leading to incomplete reconstruction in these areas.

\paragraph{Double-Sided Objects.}
Our method relies on rendering-based exploration information derived from the rendered visibility mask.  
For double-sided objects, rendering from the back side does not provide additional information, and such viewpoints are not considered informative.  
This results in incomplete reconstruction of back-side regions.  
To address this limitation, we plan to incorporate surface information into the rendering-based information gain in future work, enhancing the robustness of our exploration strategy.

\end{document}